\documentclass[journal,twoside,web]{ieeecolor}
\usepackage{generic}
\usepackage{cite}
\usepackage{amsmath,amssymb,amsfonts}
\usepackage{algorithmic}
\usepackage{graphicx}
\usepackage{textcomp}
\usepackage{soul}
\soulregister\cite7 
\soulregister\citep7
\soulregister\citet7
\soulregister\ref7
\soulregister\pageref7
\soulregister\eqref7
\usepackage{colortbl}
\usepackage{threeparttable}
\usepackage{subfigure}
\usepackage{url}
 
\def\BibTeX{{\rm B\kern-.05em{\sc i\kern-.025em b}\kern-.08em
    T\kern-.1667em\lower.7ex\hbox{E}\kern-.125emX}}
\begin{document}
\title{Direct Training via Backpropagation for Ultra-low Latency Spiking Neural Networks with Multi-threshold}
\author{Changqing Xu, \IEEEmembership{Member, IEEE}, Yi Liu and Yintang Yang, \IEEEmembership{Member, IEEE}
\thanks{“This work was supported in part by the National Natural Science Foundation of China Youth fund under Grant 62004146, by the China Postdoctoral Science Foundation funded project under Grant 2021M692498, by the Industry-University-Academy Cooperation Program of Xidian University-Chongqing IC Innovation Research Institute under Grant CQIRI-2021CXY-Z01, and by the Fundamental Research Funds for the Central Universities.”}
\thanks{Changqing Xu is with the  School of Microelectronics and Guangzhou Institute of Technology, Xidian University, Xi’an 710071, China. 
Yi Liu, and YinTang Yang,
are with the School of Microelectronics, Xidian University, Xi’an 710071, China. 
Changqing Xu is Corresponding authors (e-mail:
cqxu@xidian.edu.cn)  }
}

\maketitle

\begin{abstract}
Spiking neural networks (SNNs) can utilize spatio-temporal information and have a nature of energy efficiency which is a good alternative to deep neural networks(DNNs). The event-driven information processing makes SNNs can reduce the expensive computation of DNNs and save a lot of energy consumption. However, high training and inference latency is a limitation of the development of deeper SNNs. SNNs usually need tens or even hundreds of time steps during the training and inference process which causes not only the increase of latency but also the waste of energy consumption. To overcome this problem, we proposed a  novel training method based on backpropagation (BP) for ultra-low latency(1-2 time steps) SNN with multi-threshold. In order to increase the information capacity of each spike, we introduce the multi-threshold Leaky Integrate and Fired (LIF) model. In our proposed training method, we proposed three approximated derivative for spike activity to solve the problem of the non-differentiable issue which cause difficulties for direct training SNNs based on BP. The experimental results show that our proposed method achieves an average accuracy of $\textbf{99.56\%}$, $\textbf{93.08\%}$, and $\textbf{87.90\%}$ on MNIST, FashionMNIST, and CIFAR10, respectively with only 2 time steps. For the CIFAR10 dataset, our proposed method achieve $\textbf{1.12\%}$ accuracy improvement over the previously reported direct trained SNNs with fewer time steps.

\end{abstract}

\begin{IEEEkeywords}
Spiking neural networks, Multi-threshold, Backpropagation, Ultra low latency 
\end{IEEEkeywords}

\section{Introduction}
\label{sec:introduction}
\IEEEPARstart{S}{piking} neural networks (SNNs) is a brain-inspired neural model, which can utilize spatio-temporal information and have event-driven nature.  Unlike traditional artificial neural networks (ANNs) which consist of static and continuous-valued neuron models, SNNs process dynamic and discrete-valued spike events by more biological spiking neuron models\cite{ci1}. These characters make SNNs have potential in computational and energy efficiency on hardware neuromorphic computing systems \cite{ci2}. For instance, IBM's TrueNorth \cite{ci3} and Intel's Loihi \cite{ci4}  process a single spike with a few $pJ$ of energy. IMEC proposed a spiking neural network-based chip for radar signal processing which consumes 100 times less power than traditional implementations\cite{ci5}.
 
However, unlike ANNs which only need one time forward passes, SNNs usually require multiple time steps computation to achieve decent performance, resulting in high latency and difficulty to scale to deeper architecture. So there are two challenges for training an efficient and high-performance SNN. From the algorithmic perspective, how to make full use of rich spatial-temporal information to train SNNs is the biggest challenge of training algorithms. The non-differentiability of discrete spike events make it difficult to transmit error precisely. Second, how to minimize the time steps of SNNs without accuracy loss is the key to scaling SNNs to deeper architecture. Fewer time steps, which mean less temporal information SNNs can utilize, make it difficult to train SNNs with decent performance.

In recent years, many researchers have focused on the problem of the latency of SNNs and tried to solve the problems we mentioned above. Some researchers try to propose novel encoding methods to improve the efficiency of information representation to reduce the latency of SNNs\cite{ci2,ci6,ci7,ci8}. In \cite{ci8}, authors proposed a phase coding method to encode input spikes by the phase of a global reference clock and achieve latency reduction over the rate coding for image recognition. Compared with rate coding, the proposed method can reduce the number of time steps from hundreds to tens. In \cite{ci2}, authors proposed a time compression method to compress the temporal domain of spike training which can achieve up to 16$\times$ speedup with little accuracy loss. However, the achievable latency/spike reduction of a particular code can vary widely with network architecture and application. Furthermore, due to the decrease of the temporal information, the latency of SNNs is hard to reduce to fewer than ten time steps without accuracy loss by improving the coding methods.

Except for studying on various coding methods, there are also some researchers trying to obtain low latency SNNs from neuron model and training algorithm perspective\cite{ci11,ci9,ci10}. 
In \cite{ci11}, authors proposed a method to derivative of the postsynaptic potential (PSC) to solve the problem of non-differentiability, enabling learning targeted temporal sequences with high timing precision. The proposed method can train a decent performance SNN in five time steps, but a warm-up mechanism is applied to bring up the firing activity of the network before applying the proposed method, which makes the actual time step is larger than 5. In \cite{ci9}, authors proposed a novel BP-based method which aggregates finite differences of the loss over multiple perturbed membrane potential waveforms in the neighborhood to compute accurate error gradients. Compared with \cite{ci11}, this training method can achieve a decent performance SNNs in five time steps without extra tricks such as the warm-up mechanism. In \cite{c10}, an Iterative Initialization and Retraining method for SNNs is proposed in which an ANN is pre-trained and the weights are coped to the SNN. They explore the feasibility of directly reducing latency to one timestep by modifying the weights and thresholds. Though the proposed method can reduce the latency to one time step, it's still a kind of ANNs-to-SNNs method in which only spatial information is used during the training process. This method gives up the potential of Spatio-temporal information which causes the performance of converted SNNs to hardly outperform the corresponding ANNs.

In this paper, we proposed architecture of SNN with multi-threshold LIF models which can transmit and process more information at each time step, and the corresponding direct training method based on BP algorithm to reduce the latency requirement of SNNs. To address the issue of non-differentiability, we proposed three curves to approximate the derivative of spike activity of multi-threshold LIF models. Combining the SNN with multi-threshold LIF models and our proposed training algorithm, we can successfully train SNNs on a very short latency, e.g. two time steps. We test our SNNs framework by using the fully connected and convolution architecture on MNIST\cite{c2}, FashionMNIST\cite{c3} and CIFAR10\cite{c4} and results show that Our proposed method achieves accuracy improvement over the previously reported SNN work with lower latency. In addition, we also explore the impact of derivative approximation curves, the number of time steps, and etc. This work may help people to choose the proper parameters of our method and achieve higher performance SNNs.

\section{Approach}
\subsection{Multi-threshold Spiking Neuron Model}

\begin{figure}[!t]
\subfigure[]{
\includegraphics[width=\columnwidth]{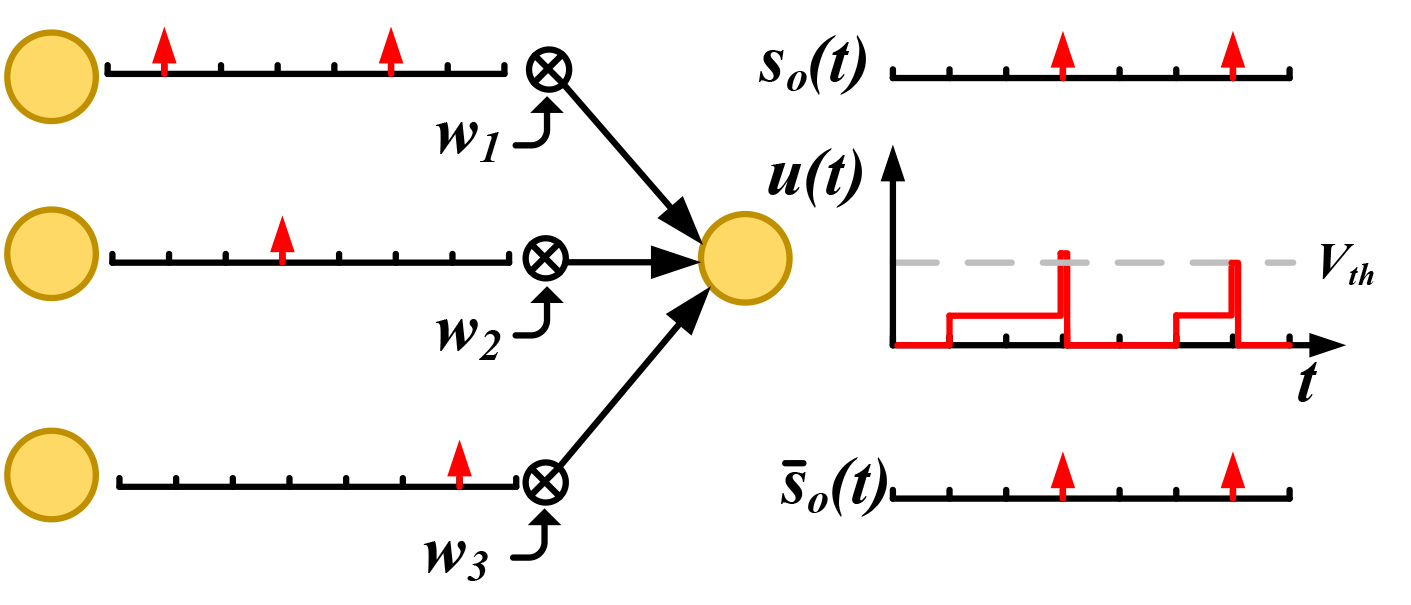}
\label{figa0}
}

\subfigure[]{
\includegraphics[width=\columnwidth]{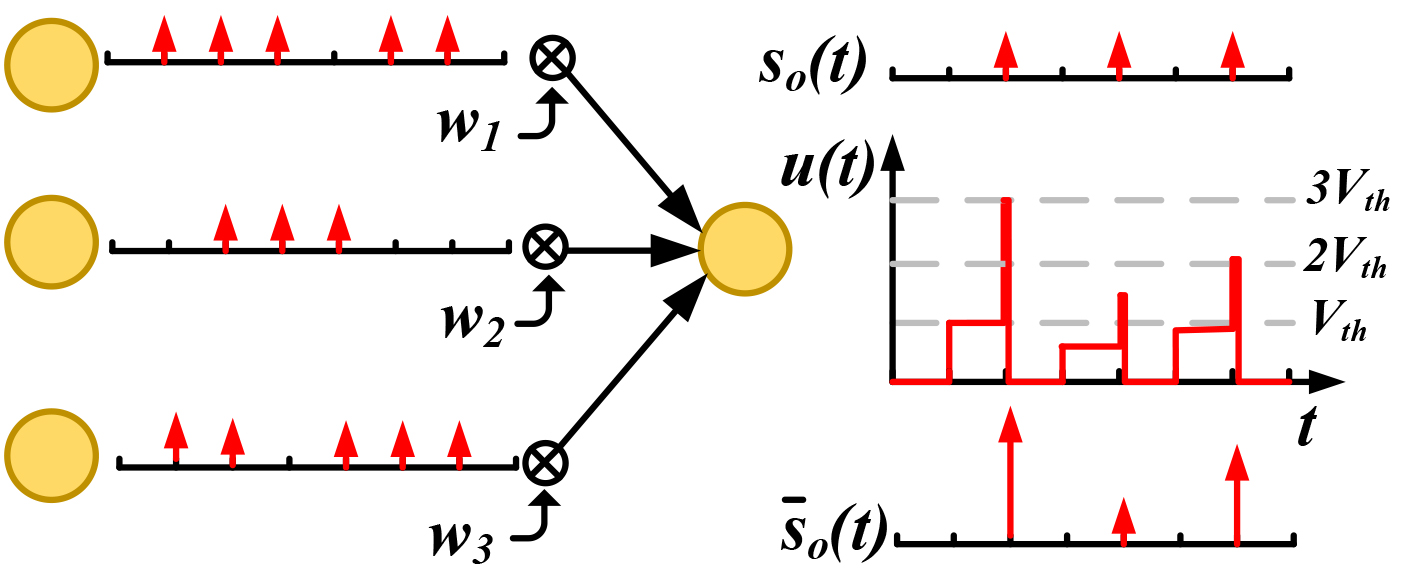}
\label{figb0}
}

\caption{Illustration of discrete LIF model and multi-threshold LIF model (a)sparse spiking input (b) dense spiking input}
\label{fig0}
\end{figure}
It is known that Leaky Integrate-and Fire (LIF) model is one of the most widely applied models to describe the neuronal dynamics in SNNs. In this paper, we introduce the LIF neuron model and synaptic model\cite{c1} adopted in our work. The neuronal membrane potential of neuron $i$ at time $t$, $u_i(t)$ is governed by
\begin{equation}
{\tau _m}\frac{{d{u_i}(t)}}{{dt}} =  - {u_i}(t) + I(t) + {u_{reset}}(t)
\label{eq1}
\end{equation}
Where $I(t)$ is the pre-synaptic input current at time $t$ and $\tau_m$ is a time constant of membrane voltage. $u_{reset}(t)$ denotes the reset function, which reduces the membrane potential by a certain amount $V_{th}$ after the neuron $i$ fires. The pre-synaptic input $I(t)$ is given by

\begin{equation}
I(t) = \sum\limits_{j = 1}^N {{\omega _{ij}}a_j(t)} 
\label{eq2}
\end{equation}

Where $N$ is the number of the pre-synaptic neurons, $\omega_{ij}$ is the presynaptic weight from the neuron $j$ in the pre-synaptic layer to the neuron $i$ in the post-synaptic layer, $a_j(t)$ is the synapse response of the pre-synaptic neuron $j$. In this paper, we apply zero-th order synaptic model and \eqref{eq2} can be simplified to

\begin{equation}
I(t) = \sum\limits_{j = 1}^N {{\omega _{ij}}s_j(t)} \label{eq3}
\end{equation}

Where $s_j(t)$ is the output spike of pre-synaptic neuron $j$ at time $t$.
Due to the discrete time steps in simulation, we apply the fixed-step Euler method to discretize \eqref{eq1} to

\begin{equation}
{u_i}[t] = (1 - \frac{1}{{{\tau _m}}}){u_i}[t - 1] + I[t] + {u_{reset}}[t]
\label{eq4}
\end{equation}

Where $u_{reset}[t]$ is equal to $-s_i[t]V_{th}$ and $s_i[t]$ is the output spike of neuron $i$. In this paper, we extend the LIF model into multi-threshold LIF model, in which the output of the neuron $i$ can be expressed by

\begin{equation}
\resizebox{.9\hsize}{!}{
${s_i}[t] = \left\{ {\begin{array}{*{20}{r}}
{0,}&{{u_i}[t] < {V_{th}}}\\
{floor(\frac{{{u_i}[t]}}{{{V_{th}}}}),}&{{V_{th}} \le {u_i}[t] < {S_{max}}{V_{th}}}\\
{{s_{\max }},}&{{u_i}[t] \ge {S_{max}}{V_{th}}}
\end{array}} \right.$
}
\label{eq5}
\end{equation}

Where $S_{max}$ is the upper limit of the output spikes and $floor()$ is the function that rounds the elements to the nearest integers towards minus infinity. Fig. \ref{fig0} shows the difference between LIF model and the multi-threshold LIF model. $s_o(t)$ is the output of LIF models and $\overline{s_o}(t)$ is the output of multi-threshold LIF models. When the input spike is sparse, the LIF model and multi-threshold LIF model can transmit the same information by generating output spikes. However, the membrane voltage may reach several times of threshold at one time step, when input spikes are dense. In this case, the LIF model will drop some information during the information transmission in the network which may cause performance loss. However, as Fig. \ref{figb0} shows that multi-threshold LIF model can keep the information by introducing a multi-threshold mechanism. In other words, the multi-threshold LIF model has more power information capacity.

\subsection{Proposed Methods}
\subsubsection{Forward Pass}

\begin{figure}[!t]
\centerline{\includegraphics[width=\columnwidth]{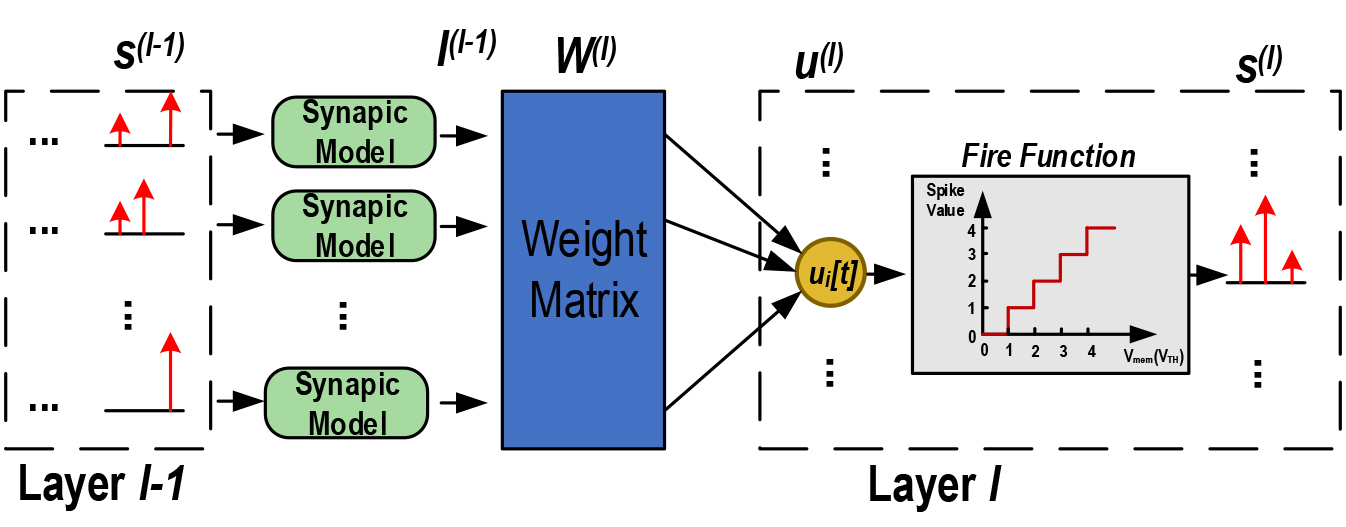}}
\caption{Forward pass of SNN.}
\label{fig1}
\end{figure}

Without loss of generality, we use two adjacent layers $l-1$ and $l$ with $N_{l-1}$ and $N_l$ neurons, respectively, to show the forward pass of a fully-connected feed forward SNN which is shown in Fig. \ref{fig1}. In the forward pass, spike train $s^{(l-1)}[t] = [s_1^{(l-1)}[t], ... ,s_{N_{l-1}}^{(l-1)}[t]]$ of the $l-1$ layer generate the pre-synaptic input $I^{(l-1)}[t] = [i^{(l-1)}_1[t], ... , i^{(l-1)}_{N_{l-1}}[t]]$ by multiplied the corresponding synaptic weight matrix $W^{(l)} = [\omega_1^{(l)}, ... , \omega_{N_l}^{(l)}]$ in which $\omega_{N_l}^{(l)} = [\omega_{1,N_l}^{(l)}; \omega_{2,N_l}^{(l)}; ... ;\omega_{N_{l-1},N_l}^{(l)}]$. The membrane potentials $u^l[t] = [u_1^l[t], ... ,u_{N_l}^l[t]]$ update based on \eqref{eq4} and trigger the output spikes of the layer $l$ neurons $s^{(l)}[t] = [s_1^{(l)}[t], ... ,s_{N_{l}}^{(l)}[t]]$ based on \eqref{eq5} when the membrane potentials exceed the threshold.

\subsubsection{Backforward Pass \label{func}}
In order to present our proposed learning algorithm, we define the following loss function $L[t_k]$ in which the mean square error for each output neuron at time step $t_k$ is to be minimized

\begin{equation}
L[{t_k}] = \frac{1}{2}{\sum\limits_{i = 0}^{{N_o}} {({y_i}[{t_k}] - {s_i}[{t_k}])} ^2}
\label{eq6}
\end{equation}

Where $N_o$ is the number of neurons in the output layer, $y_i[t_k]$ and $s_i[t_k]$ denotes the desired and the actual firing event of neurons $i$ in the output layer at time step $t_k$.

\begin{figure}[!t]
\centerline{\includegraphics[width=\columnwidth]{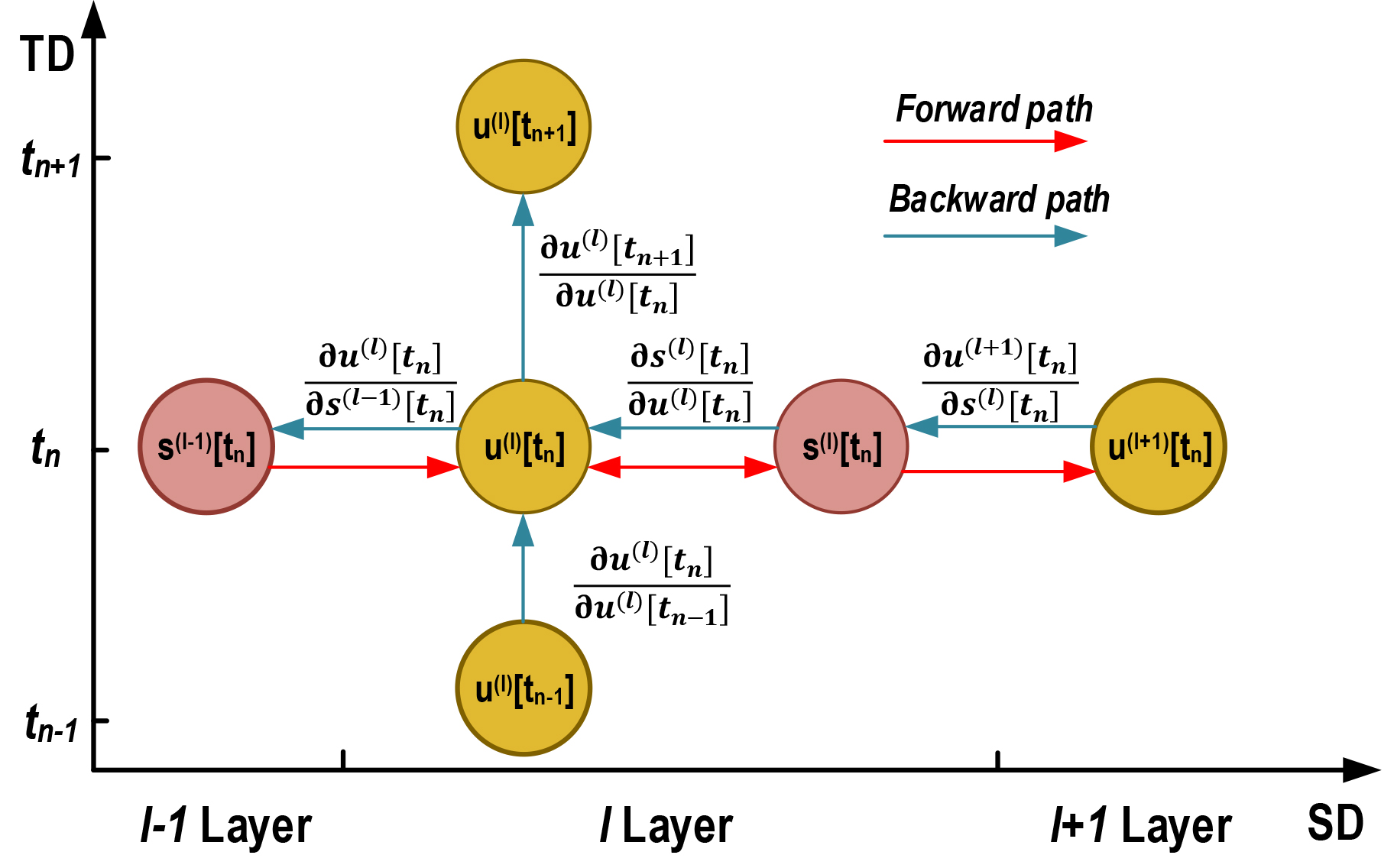}}
\caption{Error propagation in the STD.}
\label{fig2}
\end{figure}

By combining \eqref{eq1}-\eqref{eq6} together, it can be seen that loss function $L[t_k]$ is a function of synaptic weight $W$ which is required for our proposed algorithm based on gradient descent. Fig. \ref{fig2} shows the error propagation in the spatial and temporal domain(STD). The data flow of error propagation in the SD is similar to the typical BP for DNNs. Each neuron accumulates the weighted error from the upper layer and iteratively updates its parameters. In the TD, the current state of membrane voltage is dependent on its previous state, which makes it complicated to obtain the $\partial L[t_k]/\partial W$.

The goal of backforward pass is to update the synaptic weights $W$ using the error gradient $\partial L[t_k]/\partial W$ at time step $t_k$. Using the chain rule, the error gradient with the respect to the presynaptic weight $W^{(l)}$ in the layer $l$ is

\begin{equation}
\frac{{\partial L[{t_k}]}}{{\partial {W^{(l)}}}} = \frac{{\partial L[{t_k}]}}{{\partial {u^{(l)}}[{t_k}]}}\frac{{\partial {u^{(l)}}[{t_k}]}}{{\partial {W^{(l)}}}}
\label{eq7}
\end{equation}

We use $\delta ^{(l)}[{t_k}]$ to denote $\partial L[t_k]/\partial u^{(l)}[t_k]$ which is the back propagated error of layer $l$ at time $t_k$, and  \eqref{eq7} can be written as

\begin{equation}
\frac{{\partial L[{t_k}]}}{{\partial {W^{(l)}}}} = {\delta ^{(l)}}[{t_k}]\frac{{\partial {u^{(l)}}[{t_k}]}}{{\partial {W^{(l)}}}}
\label{eq8}
\end{equation}

From \eqref{eq8}, the first term $\delta ^{(l)}[{t_k}]$ can be computed from

\begin{equation}
\begin{array}{l}
{\delta ^{(l)}}[{t_k}] = \frac{{\partial {u^{(l + 1)}}[{t_k}]}}{{\partial {u^{(l)}}[{t_k}]}}\frac{{\partial L[{t_k}]}}{{\partial {u^{(l + 1)}}[{t_k}]}}\\
{\rm{            = }}\frac{{\partial {s^{(l)}}[{t_k}]}}{{\partial {u^{(l)}}[{t_k}]}}\frac{{\partial {u^{(l + 1)}}[{t_k}]}}{{\partial {s^{(l)}}[{t_k}]}}{\delta ^{(l + 1)}}[{t_k}]\\
{\rm{          }} = \frac{{\partial {s^{(l)}}[{t_k}]}}{{\partial {u^{(l)}}[{t_k}]}}{({W^{(l + 1)}})^T}{\delta ^{(l + 1)}}[{t_k}]
\end{array}
\label{eq9}
\end{equation}

For the output layer $l$, \eqref{eq9} can be computed from

\begin{equation}
\begin{array}{l}
{\delta ^{(l)}}[{t_k}] = \frac{{\partial L[{t_k}]}}{{\partial {s^{(l)}}[{t_k}]}}\frac{{\partial {s^{(l)}}[{t_k}]}}{{\partial {u^{(l)}}[{t_k}]}}\\
 = ({s^{(l)}}[{t_k}] - y[{t_k}])\frac{{\partial {s^{(l)}}[{t_k}]}}{{\partial {u^{(l)}}[{t_k}]}}
\end{array}
\label{eq10}
\end{equation}

From \eqref{eq4}, the second term $\partial u^{(l)}[t_k]/\partial W^{(l)}$ is given by

\begin{equation}
\begin{array}{l}
\frac{{\partial {u^{(l)}}[{t_k}]}}{{\partial {W^{(l)}}}} = (1 - \frac{1}{\tau_m })\frac{{\partial {u^{(l)}}[{t_{k - 1}}]}}{{\partial {W^{(l)}}}} + {s^{l - 1}}[{t_k}] - \frac{{\partial {s^{(l)}}[{t_k}]}}{{\partial W^{(l)}}}{V_{th}}\\
{\rm{ = }}(1 - \frac{1}{\tau_m })\frac{{\partial {u^{(l)}}[{t_{k - 1}}]}}{{\partial {W^{(l)}}}} + {s^{l - 1}}[{t_k}] - \frac{{\partial {s^{(l)}}[{t_k}]}}{{\partial {u^{(l)}}[{t_k}]}}\frac{{\partial {u^{(l)}}[{t_k}]}}{{\partial W^{(l)}}}{V_{th}}
\end{array}
\label{eq11}
\end{equation}

Based on \eqref{eq11}, we can obtain

\begin{equation}
\frac{{\partial {u^{(l)}}[{t_k}]}}{{\partial W^{(l)}}}{\rm{ = }}\frac{{(1 - \frac{1}{\tau_m })\frac{{\partial {u^{(l)}}[{t_{k - 1}}]}}{{\partial {W^{(l)}}}} + {s^{l - 1}}[{t_k}]}}{{1 + \frac{{\partial {s^{(l)}}[{t_k}]}}{{\partial {u^{(l)}}[{t_k}]}}{V_{th}}}}
\label{eq12}
\end{equation}

When $t_{k-1}$ is 0, \eqref{eq12} can be simplified to

\begin{equation}
\frac{{\partial {u^{(l)}}[{t_k}]}}{{\partial W^{(l)}}}{\rm{ = }}\frac{{{s^{l - 1}}[{t_k}]}}{{1 + \frac{{\partial {s^{(l)}}[{t_k}]}}{{\partial {u^{(l)}}[{t_k}]}}{V_{th}}}}
\label{eq13}
\end{equation}

As shown above, for both the output layer and hidden layers, once $\frac{{\partial {s^{(l)}}[{t_k}]}}{{\partial {u^{(l)}}[{t_k}]}}$ is known, the error can be back propagated and the gradient of each layer and the update of weights can be calculated.

\begin{figure}[!t]
\centerline{\includegraphics[width=\columnwidth]{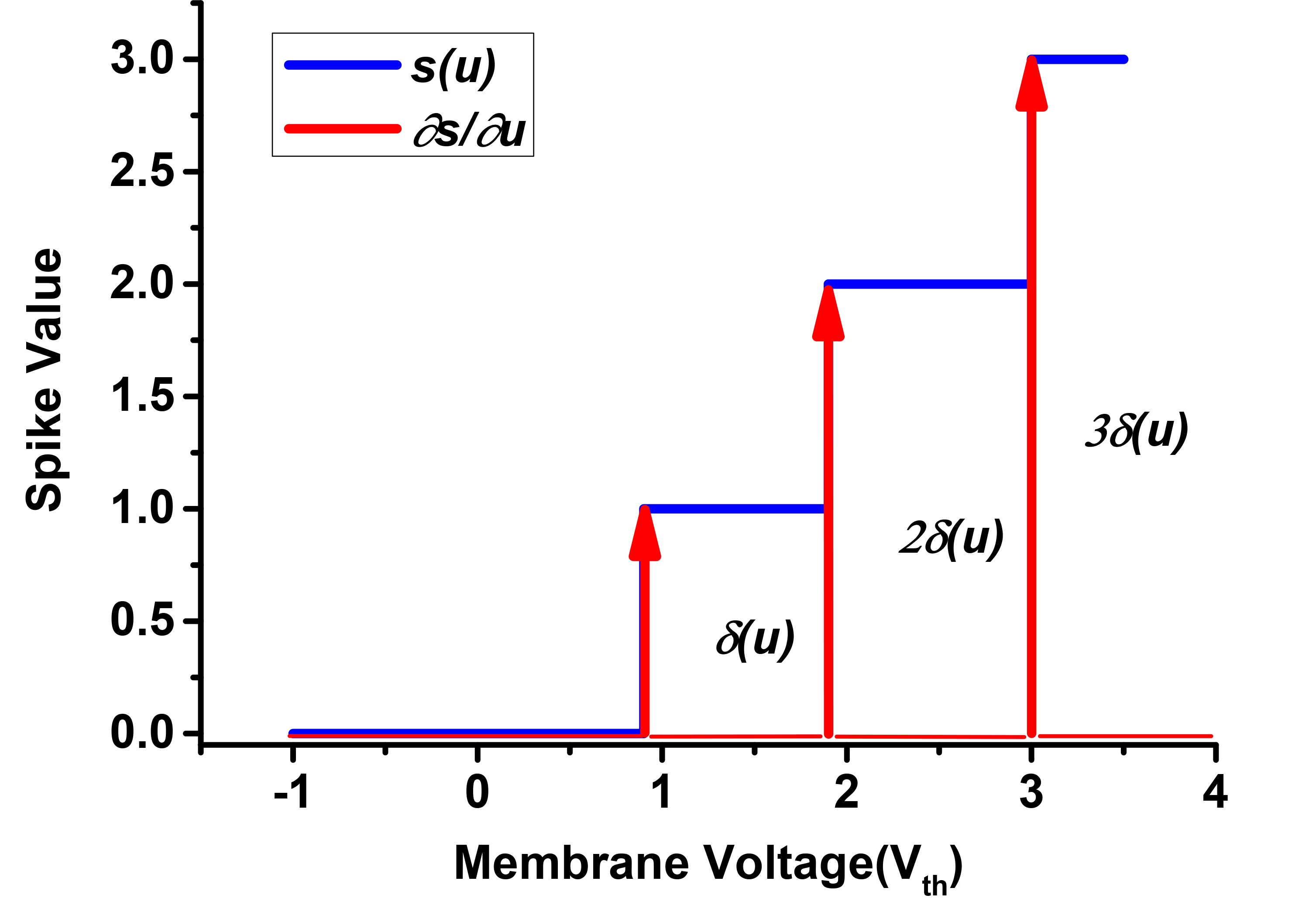}}
\caption{Multi-step activation function of the spike activity and its original derivative function.}
\label{fig3}
\end{figure}

\begin{figure}[!t]
\centerline{\includegraphics[width=\columnwidth]{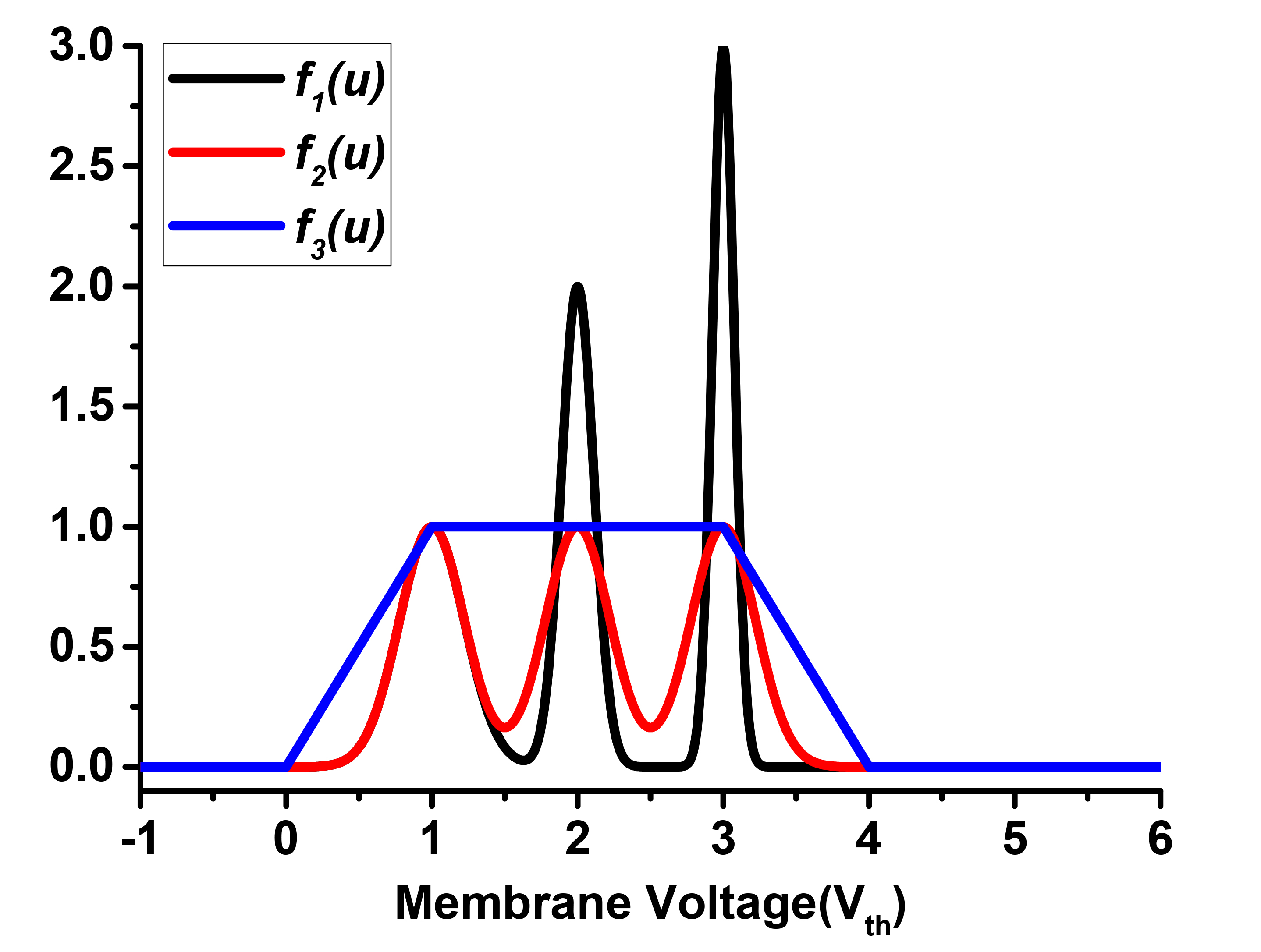}}
\caption{Three curves to approximate the derivative of spike activity.}
\label{fig4}
\end{figure}

Theoretically, $s^{(l)}[t]$ is a non-differentiable function which greatly challenges the effective learning of SNNs. Fig. \ref{fig3} shows the multi-step activation function of the spike activity and its original derivative function which is a set of Dirac function  with infinite value at $u = nV_{th}, (n>0)$ and zero value at other points. To solve this problem, we introduce three curves to approximate $\frac{{\partial {s^{(l)}}[{t_k}]}}{{\partial {u^{(l)}}[{t_k}]}}$ by $f_1$, $f_2$, and $f_3$. 

\begin{equation}
{f_1}(u) = \sum\limits_{i = 1}^{{s_{\max }}} {i{\alpha _H}{e^{ - {{(u - i{V_{th}})}^2}/({\alpha _W}/i)}}}
\label{eq14}
\end{equation}

\begin{equation}
{f_2}(u) = \sum\limits_{i = 1}^{{s_{\max }}} {{\alpha _H}{e^{ - {{(u - i{V_{th}})}^2}/{\alpha _W}}}}
\label{eq15}
\end{equation}

\begin{equation}
\resizebox{.85\hsize}{!}{
${f_3}(u) = \left\{ {\begin{array}{*{20}{r}}
{0,}&{u < 0||u \ge ({S_{\max }} + 1){V_{th}}}\\
{\frac{{{\alpha _H}u}}{{{V_{th}}}},}&{0 \le u < {V_{th}}}\\
{{\alpha _H},}&{{V_{th}} \le u < {S_{\max }}{V_{th}}}\\
{{\alpha _H}({S_{\max }} + 1 - \frac{u}{{{V_{th}}}}),}&{{S_{\max }}{V_{th}} \le u < ({S_{\max }} + 1){V_{th}}}
\end{array}} \right.$
}
\label{eq16}
\end{equation}

where $\alpha_H$ and $\alpha_W$ determine the curve shape and steep degree. $s_{max}$ is the upper limit of the output spikes.
In section \ref{sec1}, we will compare and analyze the influence on the SNNs performance with different curves and different values of parameters, such as $\alpha_H$ and $s_{max}$.
Fig. \ref{fig4} shows the curve of $f_1$, $f_2$, and $f_3$, when $s_{\max }$ is 3  and $\alpha_H$ and $\alpha_W$ are both 1.

\section{Experiments and Results}
We test our proposed SNN model and training method on three image datasets MNIST\cite{c2}, FashionMNIST\cite{c3} and CIFAR10\cite{c4} with different sizes and structures of SNNs. And we compared our training method with several previously reported state-of-the-art results with the same or similar networks including different SNNs trained by BP-based methods, converted SNNs, and traditional ANNs.

\subsection{Experiment Settings}
All reported experiments below are conducted on an NVIDIA Tesla V100 GPU. The implementation of our proposed method is on the Pytorch framework\cite{c5}. The experimented SNNs are based on the multi-threshold LIF model described in \eqref{eq3}-\eqref{eq5}. The simulation step size is set to $1 ms$. Function $f_2$ is applied to approximate the derivative of spike activity. Only two time steps are used to demonstrate the proposed ultra low-latency spiking neural network. No refractory period is used. Adam \cite{c6} is applied as the optimizer. If not otherwise specified, the accuracy in this paper refers to the mean of the accuracy obtained by repeating the experiments five times.

\subsection{Parameter Initialization}
The initialization of parameters, such as the weights, threshold, time constant of membrane voltage, and other parameters, directly affect the convergence speed and stability of the whole network. We should simultaneously make sure enough spikes transmit information between neural network layers and avoid too many spikes that reduce the neuronal selectivity. As it is known that weight and threshold are two key factors for the LIF model in the forward pass\cite{c7}. In this paper, we use a fixed threshold in each neuron for simplification and initial the weight $W^{(l)}$ parameters sampling from the normal distribution. 

\begin{equation}
W^{(l)}{\rm{\sim[}}\frac{{{V_{th}}}}{{{N_{l-1}}}}{\rm{,0}}{\rm{.5]}}
\label{eq17}
\end{equation}

where $V_{th}$ is the threshold of membrane voltage, $N_{l-1}$ is the number of neurons of pre-layer. The set of other parameters is presented in Table \ref{table1}. In addition, we do not apply complex skill, such as error normalization\cite{c8}, dropout, weight regularization\cite{c9}, warm-up mechanism \cite{c10}, etc. In the preprocessing step, the dataset is normalized and only random cropping is applied for data augmentation. All testing accuracy is reported after training 200 epochs in our experiments.

\begin{table}[t]
\caption{Parameters setting}
\label{table1}
\setlength{\tabcolsep}{3pt}
\centering
\arrayrulecolor{black}
\begin{tabular}{lll} 
\hline
Parameters         & Description                               & Value                 \\ 
\hline
\textit{$\tau_m$}        & Time constant of membrane voltage         & 10 ms                 \\
\textit{$V_{th}$}       & Threshold                                 & 10 mV                 \\
\textit{$\alpha _m$}        & Derivative approximation parameters       & 1                     \\
\textit{$\alpha _W$}        & Derivative approximation parameters       & 20                    \\
\textit{$S_{max}$}      & Upper limit of output spikes              & 15                    \\
\textit{$N_{Batch}$}    & Batch Size                                & 128                   \\
\textit{$\eta$}         & \begin{tabular}[c]{@{}l@{}}Learning rate\\(MNIST/FashionMNIST/CIFAR10)\end{tabular} & 0.005, 0.005, 0.0005  \\
\textit{$\beta _1, \beta _2, \lambda$} & Adam parameters                           & 0.9, 0.999, $1-10^{-8}$    \\
\hline
\end{tabular}
\arrayrulecolor{black}
\end{table}

\subsection{Dataset Experiments}
\subsubsection{MNIST}
The MNIST dataset of handwritten digits \cite{c2} consists of a training set with 60,000 labeled hand-written digits, and a test set of 10,000 labeled hand-written digits, each of which is a $28\times28$ grayscale image. Each pixel value of images is converted into a real-valued input current like \cite{c10}. 

For MNIST dataset \cite{c2}, we compare several similar spiking Multi-Layer Perceptron (MLP), each of which consists of one to three hidden layers and several spiking Convolutional Neural Networks (CNNs), each of which consists of two convolutional layers, two pooling layers, and a dense layer, respectively. Table \ref{table2} shows the spiking MLP trained by our method can achieve $99.15\%$ which outperforms other reported results and obtains a large reduction of time step count.
Table \ref{table3} shows the test accuracies of the spiking CNNs trained by our proposed method and other reported algorithms except for the method in \cite{c15} whose accuracy is only slightly higher by $0.01\%$. However, our proposed method obtains up to $99.56\%$ in only two time steps which is much less than 400 time steps \cite{c15} needs. 

\begin{table}
\caption{Comparison with similar spiking MLP on MNIST}
\label{table2}
\setlength{\tabcolsep}{3pt}
\centering
\arrayrulecolor{black}
\begin{tabular}{llll} 
\hline
Methods         & Network          & Time steps & Accuracy  \\ 
\hline
Converted SNN\cite{c11}*         & 784-1200-1200-10 & 20         & 98.64\%   \\
STDP\cite{c12}          & 784-6400-10      & 350        & 95.00\%   \\
BP\cite{c9}          & 784-800-10       & 200-1000   & 98.71\%   \\
STBP\cite{c7}           & 784-800-10       & 50-300     & 98.89\%   \\
Proposed Method & 784-800-10       & 2          & 99.15\%   \\
\hline
\end{tabular}
\arrayrulecolor{black}
\begin{tablenotes}
\item[1]
* means their model is converted from the pre-trained ANN model
\end{tablenotes}
\end{table}

\begin{table}
\caption{Comparison with similar spiking CNN on MNIST}
\label{table3}
\setlength{\tabcolsep}{3pt}
\centering
\arrayrulecolor{black}
\begin{tabular}{llll} 
\hline
Methods         & Network             & Time steps & Accuracy  \\ 
\hline
SLAYER\cite{c13}               & 12C5-P2-64C5-p2     & 300       & 99.36\%   \\
HM2BP\cite{c14}               & 15C5-P2-40C5-P2-300 & 400       & 99.42\%   \\
ST-RSBP\cite{c15}               & 15C5-P2-40C5-P2-300 & 400       & 99.57\%   \\
TSSL-BP\cite{ci11}               & 15C5-P2-40C5-P2-300 & 5         & 99.47\%   \\
Proposed Method & 15C5-P2-40C5-P2-300 & 2         & 99.56\%   \\
\hline
\end{tabular}
\arrayrulecolor{black}
\begin{tablenotes}
\item[1] 20C5 represents convolution layer with 20 of the 5 $\times$ 5 filters. P2 represents pooling layer with 2 $\times$ 2 filters.
\end{tablenotes}
\end{table}

\subsubsection{FashionMNIST}
The FashionMNIST dataset of clothing items contains 60,000 labeled training images and 10,000 labeled testing images, each of which is also $28\times28$ grayscale images like MNIST. Compared with MNIST, FashionMNIST is a more challenging dataset that can serve as a direct drop-in replacement for the original MNIST dataset. We compare trained spiking MLPs and CNNs on FashionMNIST. In Table \ref{table4}, we compared our proposed method with other methods on the same architecture of two hidden layers. Our proposed method obtains $91.08\%$ test accuracy, outperforming the TSSL-BP method \cite{ci11}, which is the best previously reported algorithm for SNNs as we know. 
In addition, compared with \cite{ci11}, our method reduces the training time steps from 5 to 2, further. Table \ref{table5} compares CNNs with similar architecture. Our method achieves $93.08\%$ in two time steps which shows it outperforms other methods and reduces training time steps, noticeably.  

\begin{table}
\caption{Comparison with spiking MLP on FashionMNIST}
\label{table4}
\setlength{\tabcolsep}{3pt}
\centering
\arrayrulecolor{black}
\begin{tabular}{llll} 
\hline
Methods         & Network        & Time steps & Accuracy  \\ 
\hline
ANN\cite{c15}*               & 784-512-512-10 & ~          & 89.01\%   \\
HM2BP\cite{c17}               & 784-400-400-10 & 400        & 88.99\%   \\
ST-RSBP\cite{c15}               & 784-400-400-10 & 400        & 90.13\%   \\
TSSL-BP\cite{ci11}               & 784-400-400-10 & 5          & 90.19\%   \\
Proposed Method & 784-400-400-10 & 2          & 91.08\%   \\
\hline
\end{tabular}
\arrayrulecolor{black}
\begin{tablenotes}
\item[1]
* means their model is the non-spiking ANN trained by a standard BP method
\end{tablenotes}
\end{table}

\begin{table}
\caption{Comparison with spiking CNN on FashionMNIST}
\label{table5}
\setlength{\tabcolsep}{3pt}
\centering
\arrayrulecolor{black}
\begin{tabular}{llll} 
\hline
Methods         & Network              & Time steps & Accuracy  \\ 
\hline
ANN$^{*1}$~          & 32C5-P2-64C5-P2-1024 & ~          & 91.60\%   \\
TSSL-BP         & 32C5-P2-64C5-P2-1024 & 5          & 92.45\%   \\
Converted SNN$^{*2}$ & 16C5-P2-64C5-P2-1024 & 200        & 92.62\%   \\
Proposed Method & 32C5-P2-64C5-P2-1024 & 2          & 93.08\%   \\
\hline
\end{tabular}
\arrayrulecolor{black}
\begin{tablenotes}
\item[1]
 *1 means their model is the non-spiking ANN trained by a standard BP method
 \item[2]
 *2 means their model is converted from the pre-trained ANN model
\end{tablenotes}
\end{table}

\subsubsection{CIFAR10 \label{CIFAR10}}
To validate our method, we apply a deeper CNN which contains five convolutional layers, two pooling layers and two dense layers on the more challenging dataset of CIFAR10 \cite{c4}. Different from MNIST and FashionMNIST, CIFAR10 is a subset of the 80 million tiny images dataset and consists of 60,000 32x32 color images containing one of 10 object classes, with 6000 images per class. In addition, it is hard to scale to deeper networks for traditional SNNs which need long training latency. To the best of our knowledge, only a few works report direct training of SNNs on CIFAR10 which we list in table \ref{table6}. Our proposed method obtains $88.17\%$ accuracy with a mean of $87.90\%$. Compared with TSSL-BP\cite{ci11}, our method achieves $1.12\%$ average testing accuracy improvement with fewer time steps and fewer additional optimization and augmentation technologies.

\begin{table}
\caption{Comparison with spiking CNN on CIFAR10}
\label{table6}
\setlength{\tabcolsep}{3pt}
\centering
\arrayrulecolor{black}
\begin{tabular}{llll} 
\hline
Methods          & Skills                                                                                           & Time steps & Accuracy  \\ 
\hline
ANN\cite{c20}$^{*1}$~           & Random cropping                                                                                  & ~          & 83.72\%   \\ 
Converted SNN\cite{c20}$^{*2}$~ & Random cropping                                                                                  & ~          & 83.52\%   \\
STBP\cite{c21}             & \begin{tabular}[c]{@{}l@{}}Neuron normalization,\\dropout, and\\population decoding\end{tabular} & 8          & 85.24\%   \\
TSSL-BP\cite{ci11}          & \begin{tabular}[c]{@{}l@{}}Random cropping \\and horizontal flipping\end{tabular}                & 5          & 86.78\%   \\
Proposed Method  & Random cropping                                                                                  & 2          & 87.90\%   \\
\hline
\end{tabular}
\arrayrulecolor{black}
\begin{tablenotes}
\item[1]
 *1 means their model is the non-spiking ANN trained by a standard BP method
 \item[2]
 *2 means their model is converted from the pre-trained ANN model
 \item[3]
 The network structure is 96C3-256C3-P2-384C3-P2-384C3-256C3-1024-1024.
\end{tablenotes}
\end{table}

\subsection{Performance analysis \label{sec1}}

\subsubsection{The impact of derivative approximation curves}
\begin{figure}[!t]
\centerline{\includegraphics[width=\columnwidth]{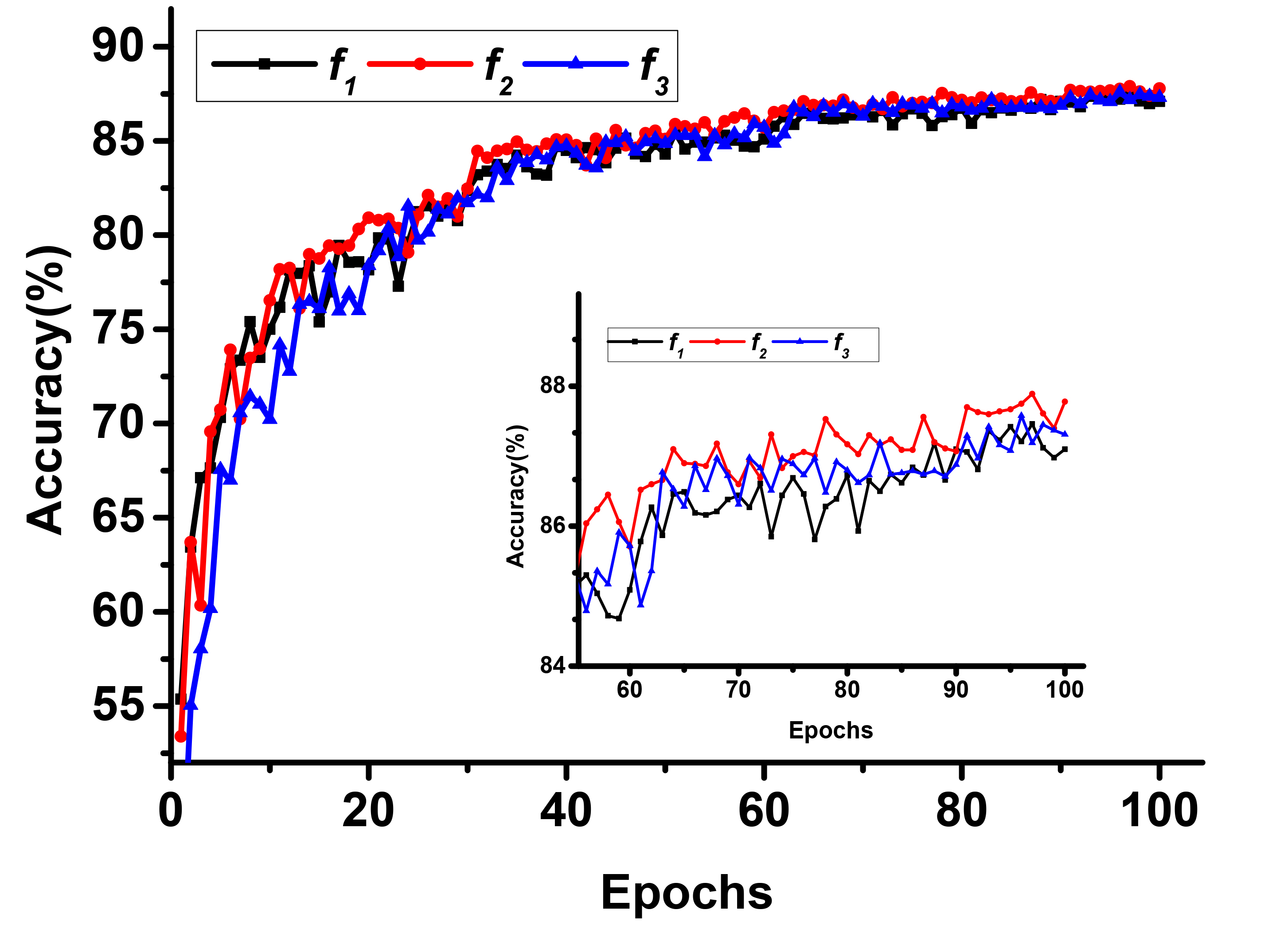}}
\caption{Comparison of different derivation approximation curves.}
\label{fig5}
\end{figure}

In section \ref{func}, we introduce three different curves to approximate the spike activity. In this section, we try to analyze the impact of different curves on the performance of the trained network. The experiments are conducted on the CIFAR10 dataset, and the network structure is 96C3-256C3-P2-384C3-P2-384C3-256C3-1024-1024, which is the same as the section \ref{CIFAR10} described. The parameter of curves is the same as table \ref{table1} lists. As Fig. \ref{fig5} shows that three curves present similar performance. Because the value of curve 1 may be larger than 1, which may cause gradient exploding problem in deep networks, curve 2 and 3 will be a better choice. Compared with curve 3, curve 2 have a slight improvement. So we chose curve 2 to approximate the spike activity in this paper.
 
\begin{figure}[!t]
\centerline{\includegraphics[width=\columnwidth]{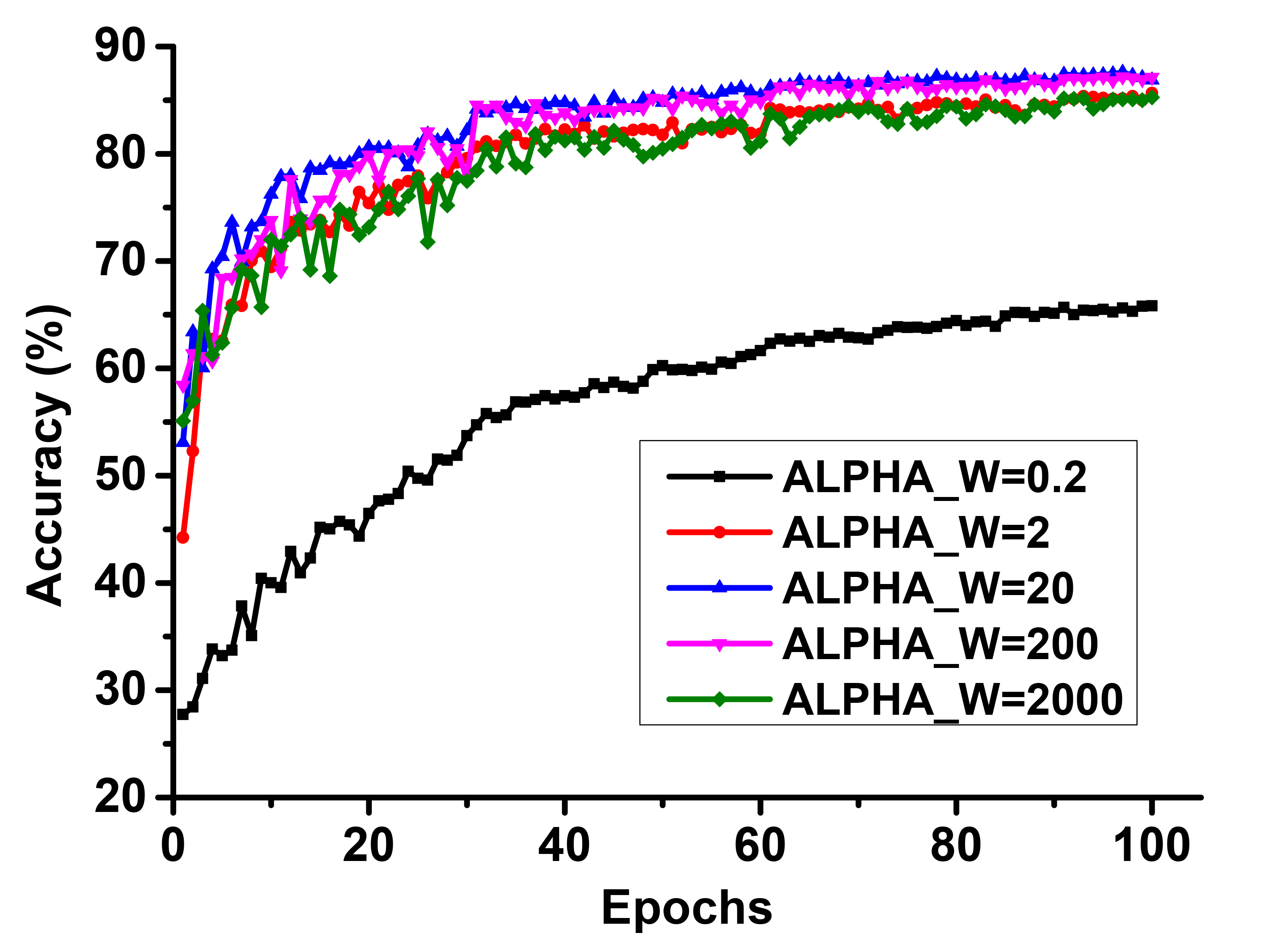}}
\caption{Impact of different widths of derivation approximation curves.}
\label{fig6}
\end{figure}

Furthermore, we chose curve 2 to explore the impact of parameters of curves. The main parameters of the derivative approximation curves are $\alpha _W$ and $\alpha _H$. Due to the gradient exploding problem we mentioned above, we fix the value of  $\alpha _H$ to 1 and focus on the impact of different widths of the curve which is controlled by $\alpha _W$. $\alpha _W$ is set to 0.2, 2, 20, 200, 2000, and two time step is applied in the experiments. The corresponding testing accuracy after 100 epochs is shown in Fig. \ref{fig6}. Too small $\alpha _W$ will cause worse performance. When $\alpha _W$ is larger than 20, the testing accuracy has a little decrease. The reason that the shape of derivative approximation curves is not very sensitive to the accuracy is the main function of these curves is to capture the nonlinear nature\cite{c7}.  

\subsubsection{The impact of $S_{max}$}
\begin{figure}[!t]
\centerline{\includegraphics[width=\columnwidth]{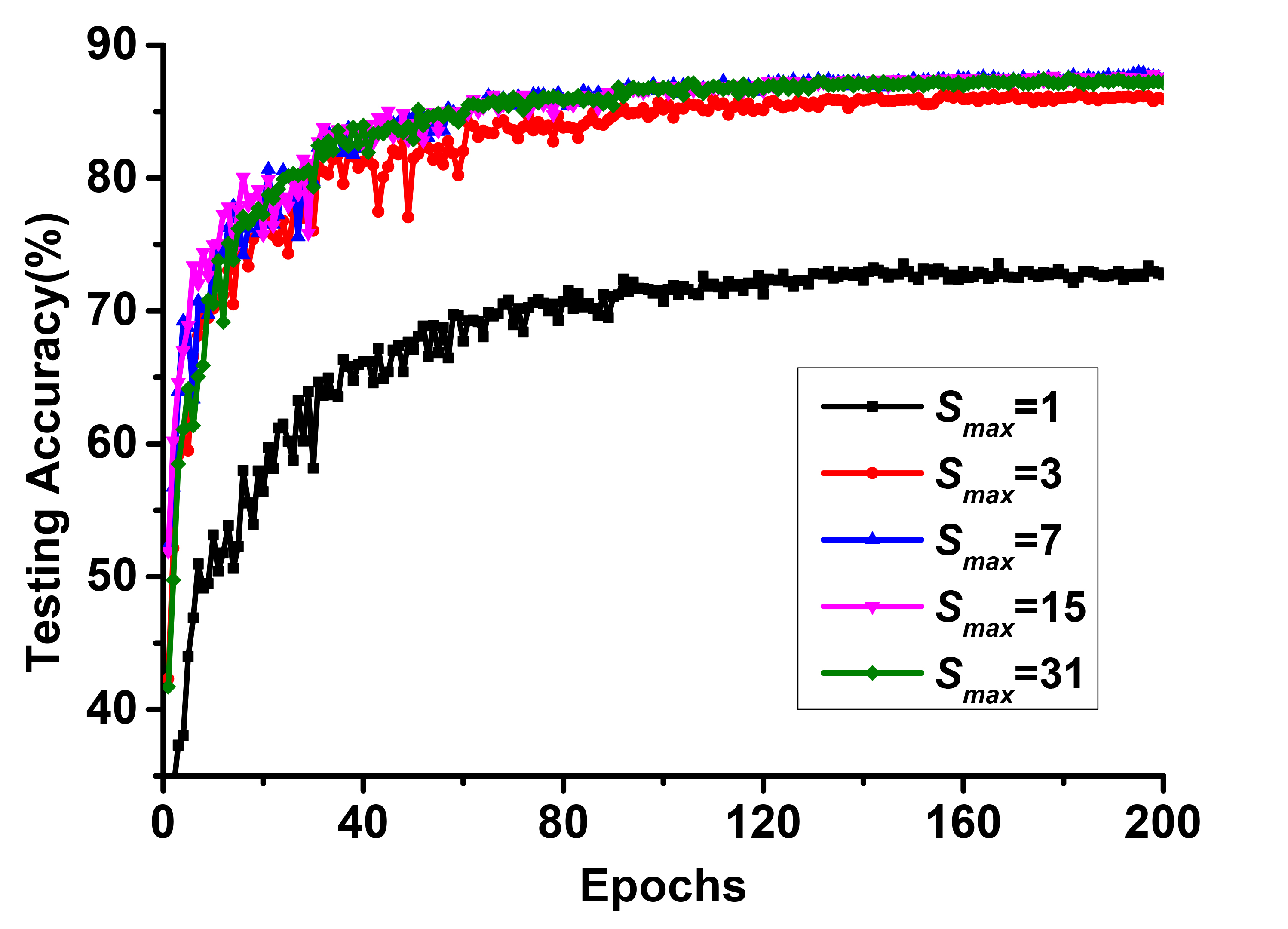}}
\caption{Impact of $S_{max}$ on Testing accuracy}
\label{fig7}
\end{figure}

$S_{max}$ is the upper limit of the output spikes which directly affect the capacity of information at each time step. In this section, we study the impact of $S_{max}$ on the accuracy. Curve 2 is applied and the same experiment configuration is set. To shield the influence of the temporal domain, the training time step is set to 1. As Fig. \ref{fig7} shows that the network still achieves a reasonable accuracy even when only a one-time step is applied if the $S_{max}$ is large enough to make spikes have enough information capacity. The testing accuracy reaches $87.82\%$ when $S_{max}$ is 15. And the testing accuracy will be saturated when $S_{max}$ is larger than 7. As we know that SNN can utilize the information of temporal domain and spatial domain, the information of each spike can be treated as spatial domain information and the position of spikes in the spike train carries temporal information. When the value of $S_{max}$ makes a spike have enough capacity of information to transmit all spatial information in the SNN, the increase of $S_{max}$ won't improve the accuracy.



\subsubsection{The impact of length of spike train}
\begin{figure}[!t]
{
\subfigure[]{
\includegraphics[width=\columnwidth]{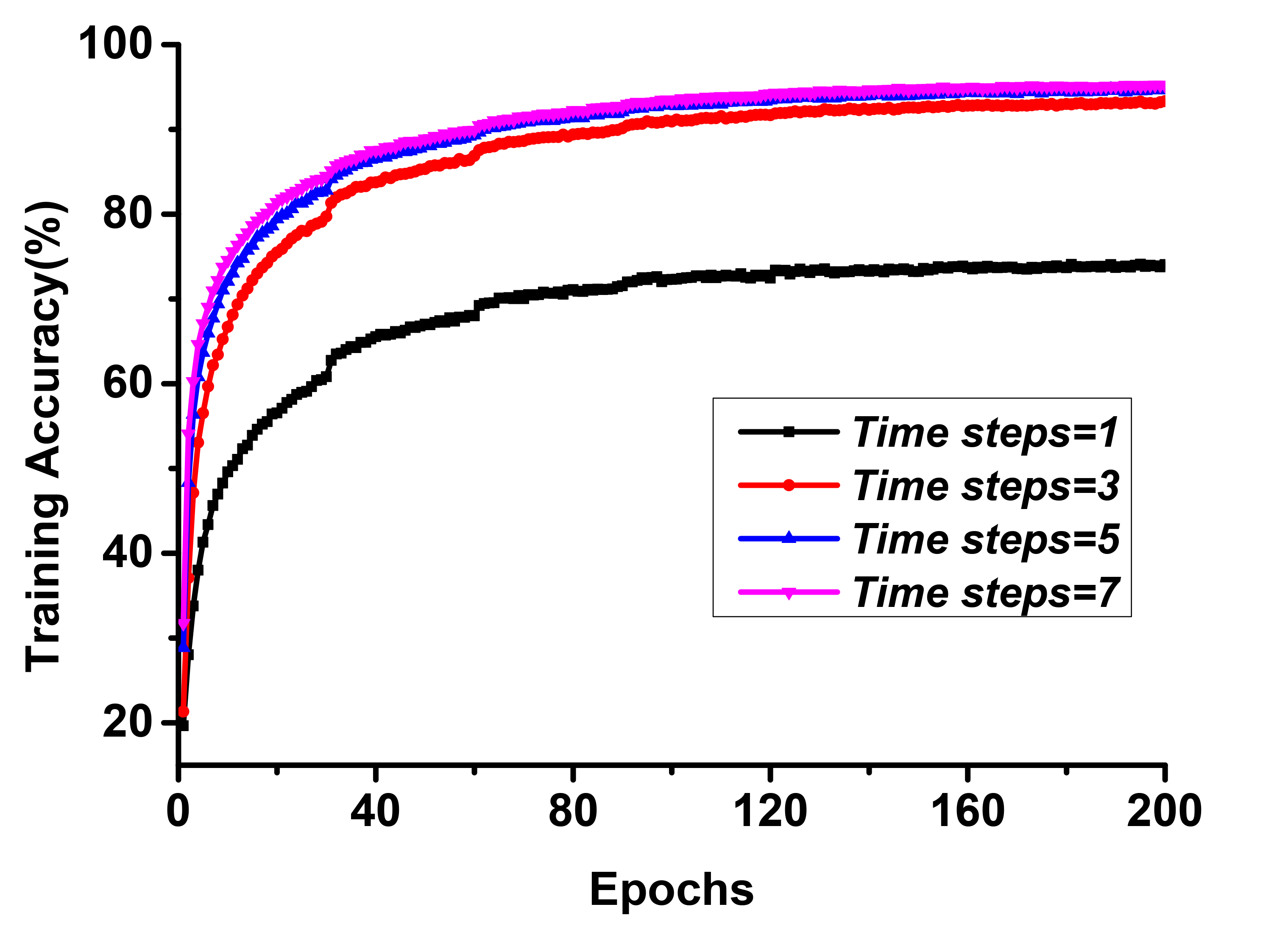}
\label{figa8}
}

\subfigure[]{
\includegraphics[width=\columnwidth]{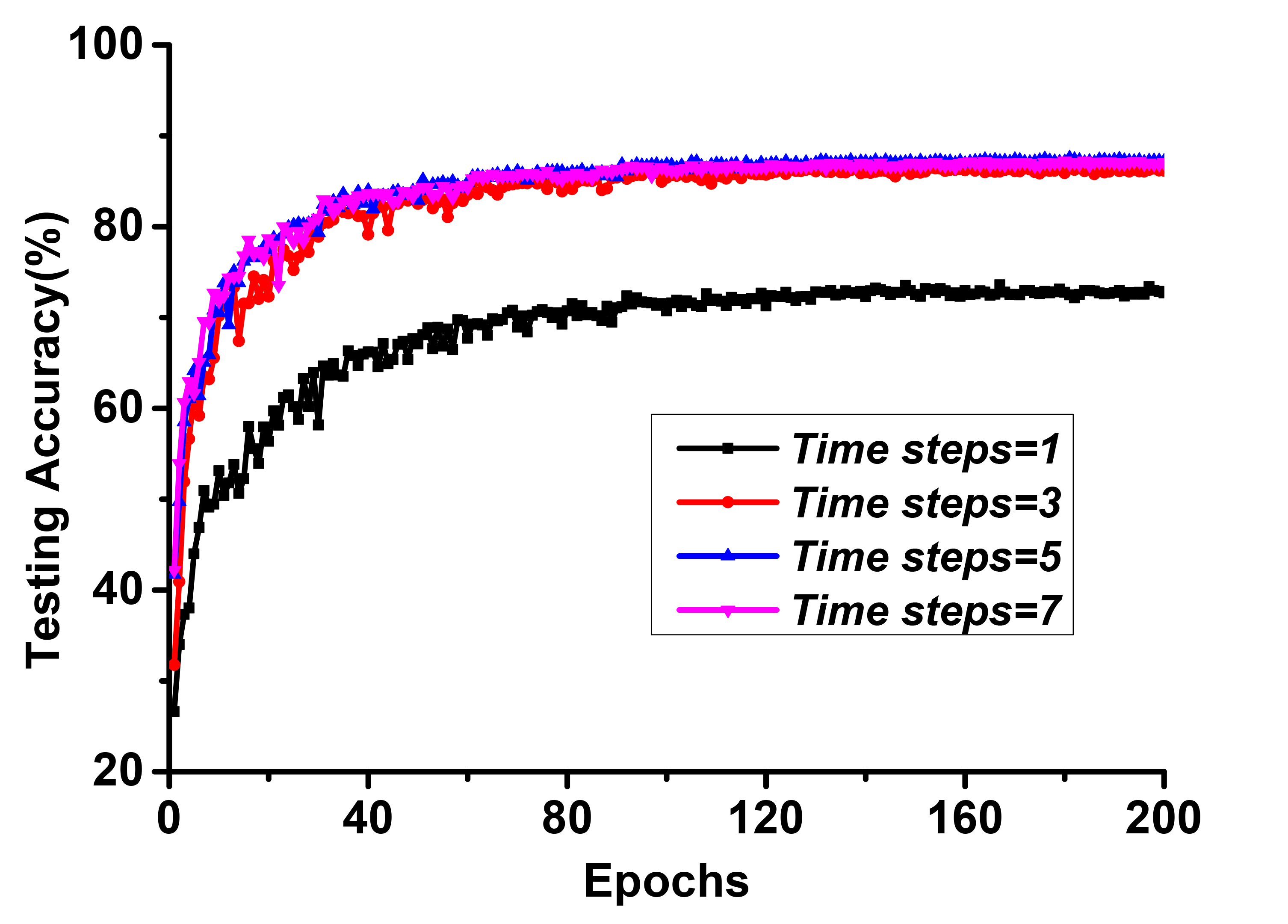}
\label{figb8}
}
\caption{Impact of the length of time steps (a) Training accuracy (b) Testing accuracy }
\label{fig8}
}
\end{figure}

For SNN, a longer spike train may carry more temporal information, theoretically. In this section, we mainly study the impact of the length of spike trains on performance. Curve 2, whose $S_{max} $ is 1, is applied and the same experiment configuration is set. The number of time steps is used to measure the length of spike trains.
Fig. \ref{fig8} shows the testing and training accuracy for different time steps. Compared with SNN with one time step which contains no temporal information, SNN with multi-time steps can improve the performance of SNN. Compared with SNN with one-time step, SNN with 5 time steps improve the testing accuracy from $73.60\%$ to $87.55\%$.  
As Fig. \ref{fig8} shows, the testing accuracy is no longer improved when time steps are larger than 3. However, the training accuracy is still improved with the increase of the time steps. The reason for that is the increase in the length of spike trains improves the capacity of temporal information which can improve the performance. However, temporal information is very sensitive which may reduce the robustness of the trained networks. Too long spike train may make network overfitting if other training skills are not applied.

\section{Conclusion}
In this paper, we presented a novel training method based on backpropagation for ultra-low latency spiking neural networks. The LIF model with multi-threshold is introduced to make the SNN carry more spatial information in each time step. We also proposed three approximated derivative curves to address the non-differentiable problem of multi-threshold spike activity and proposed a training method based on backpropagation.
We demonstrate the state-of-the-art performances in comparison with the SNN BP method, converted SNNs, and even traditional ANNs on the MNIST, FashionMNIST, and CIFAR10 datasets. Experimental results show that our proposed method outperforms other SNN BP methods with the same or similar network architecture on no matter accuracy and latency. In short, we tried to propose a method to address the high training and inference latency issue of SNNs and presents an idea of training an SNN with the ultra-low possible latency, directly



\section*{Acknowledgment}

This work was supported in part by the National Natural Science Foundation of China under Grant 62004146, by the China Postdoctoral Science Foundation funded project under Grant 2021M692498, by the Fundamental Research Funds for the Central Universities, by the Industry-University-Academy Cooperation Program of Xidian University-Chongqing IC Innovation Research Institute under Grant CQIRI-2021CXY-Z01.

\begin{IEEEbiography}[{\includegraphics[width=1in,height=1.25in,clip,keepaspectratio]{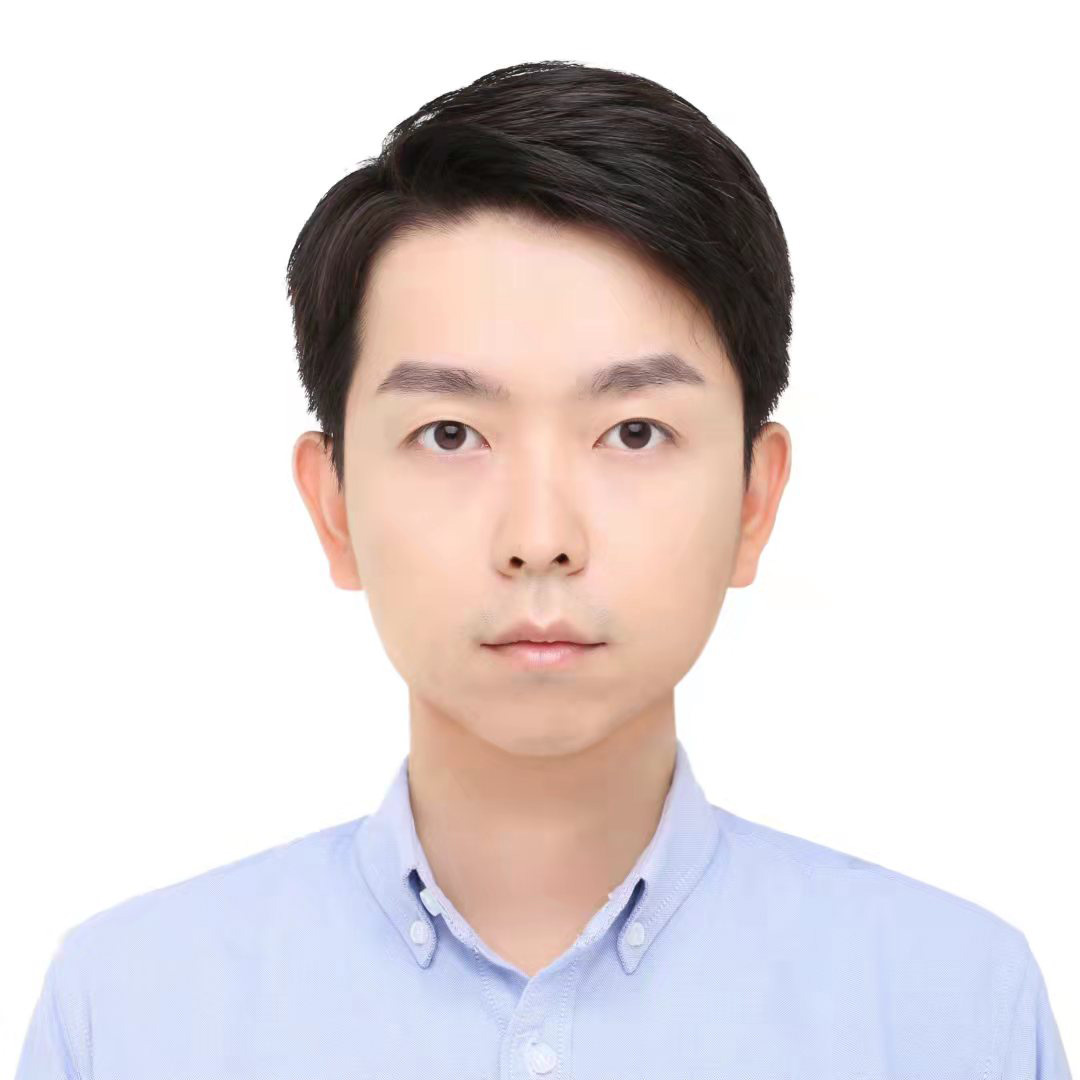}}]{Changqing Xu}(Member, IEEE) was born in Jiaozuo, Henan, China. He received the Ph. D. degree in microelectronics and solid state electronics from Xidian University, Xi’an, China, in 2019. He is an associate professor at Xidian University, Xi’an, China. His research interests are in the areas of algorithm and architecture of SNN, design and optimization of NoC and technology of LVSI simulation. He has published more than ten papers in IEEE Transactions on Computer-Aided Design of Integrated Circuits and Systems,  IEEE Transactions on Electron Devices, Frontiers in neuroscience, SCIENCE CHINA Information Sciences, Microelectronics journal and etc.
\end{IEEEbiography}

\begin{IEEEbiography}[{\includegraphics[width=1in,height=1.25in,clip,keepaspectratio]{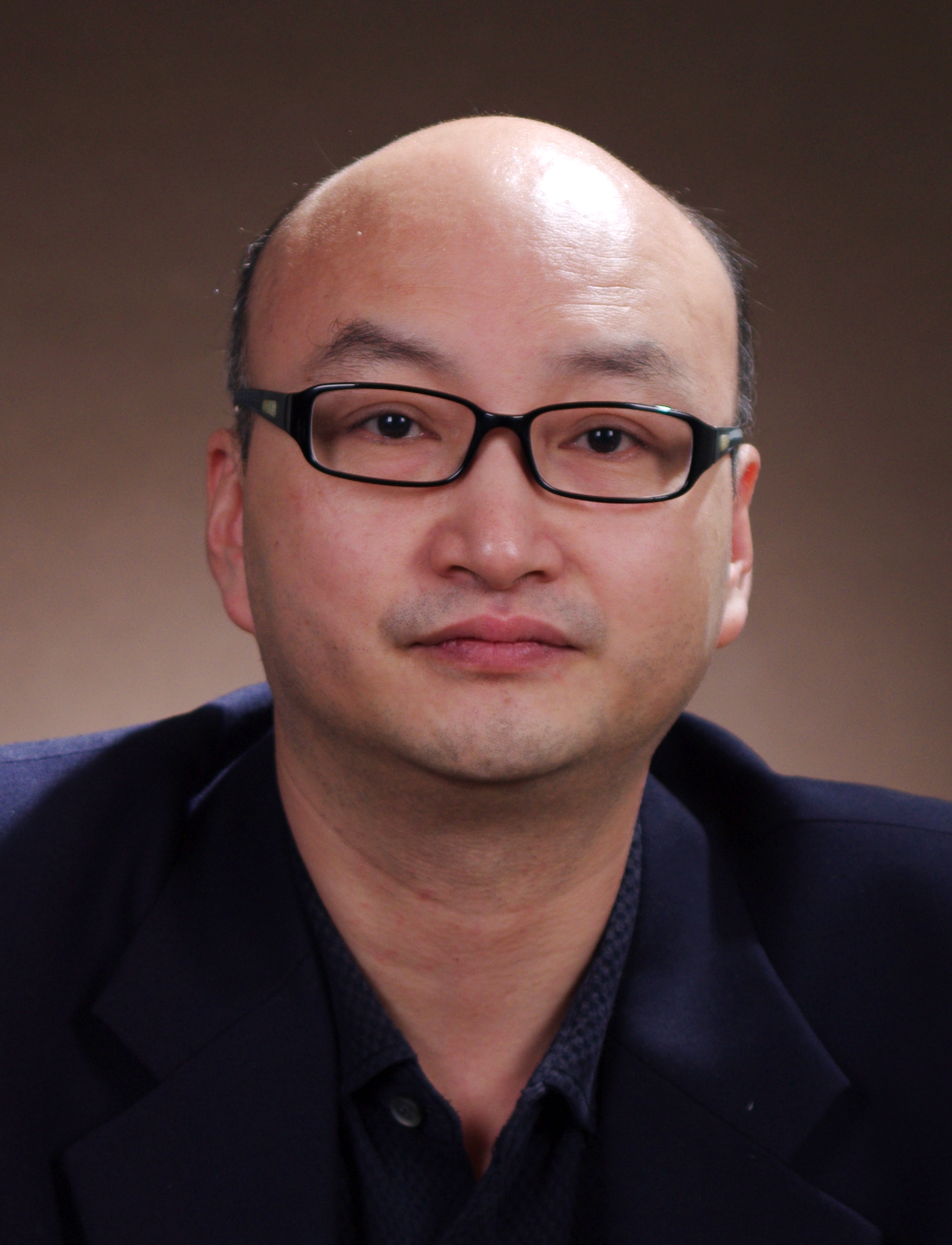}}]{Yi Liu} received the Ph. D. degree in microelectronics and solid state electronics from Xidian University, Xi’an, China, in 2010. He has been a Professor with the School of Microelectronics, Xidian University, Xi’an, China, since 2013. His research interests are in the areas of low-power coding and low-swing technology, and radiation effect simulation.
\end{IEEEbiography}

\begin{IEEEbiography}[{\includegraphics[width=1in,height=1.25in,clip,keepaspectratio]{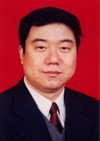}}]{YinTang Yang} (Member, IEEE) was born in Hebei, China, in 1962. He received the B.S. and M.S. degrees in microelectronics and solid state electronics from Xidian University, Xi’an, China, in 1982 and 1984, respectively, and the Ph.D. degree in electronic science and technology from Xi’an Jiaotong University, Xi’an. He has been a Professor with the School of Microelectronics, Xidian University, since 1997. 
\end{IEEEbiography}

\end{document}